\pdfoutput=1
\documentclass[11pt]{article}

\usepackage[dvipsnames]{xcolor}
\usepackage{acl}
\usepackage{times}
\usepackage{latexsym}
\usepackage[normalem]{ulem}
\usepackage{siunitx}
\usepackage{multirow}
\usepackage{booktabs}       % professional-quality tables
\usepackage{graphicx}
\usepackage{hhline}
\usepackage{arydshln}
\usepackage{xspace}

\usepackage{amsmath,amsfonts,bm}

\def\eqref#1{equation~\ref{#1}}
\def\1{\bm{1}}

\DeclareMathAlphabet{\mathsfit}{\encodingdefault}{\sfdefault}{m}{sl}
\SetMathAlphabet{\mathsfit}{bold}{\encodingdefault}{\sfdefault}{bx}{n}

\usepackage[T1]{fontenc}
\usepackage[utf8]{inputenc}

\usepackage{microtype}
\usepackage{amsmath}
\usepackage{amsfonts}
\usepackage{amssymb}
\usepackage{verbatim}
\usepackage{bm} %For bold-facing in the math mode
\usepackage{bbm}
\usepackage{placeins}
\usepackage{float}

\newcommand{\bertbase}{{$\mbox{BERT}_{Base}$}\xspace}
\newcommand{\bertlarge}{{$\mbox{BERT}_{Large}$}\xspace}
\newcommand{\encodername}{{$\mbox{MiniLM}$}\xspace}
\newcommand{\singletask}{{\mbox{Single-Task}}\xspace}
\newcommand{\densemtl}{{\mbox{MT-Dense}}\xspace}
\newcommand{\sparsemtl}{{\mbox{MT-Switch}}\xspace}
\newcommand{\msparsemtl}{{\mbox{MT-TaG}}\xspace}
\newcommand{\sixtasks}{{\mbox{C-GLUE}}\xspace}
\newcommand{\eighttasks}{{\mbox{GLUE}}\xspace}
\newcommand{\sixteentasks}{{\mbox{GLUE\texttt{++}}}\xspace}

\newcommand{\sg}[1]{{\color{red}\small{ SG: #1}}}

\usepackage{tikz}
\newcommand*\circled[1]{\tikz[baseline=(char.base)]{
            \node[shape=circle,fill,inner sep=0.5pt] (char) {\textcolor{white}{#1}};}}

\title{Sparsely Activated Mixture-of-Experts are Robust Multi-Task Learners}

\author{Shashank Gupta$^\spadesuit$, Subhabrata Mukherjee$^\spadesuit$,  Krishan Subudhi$^\clubsuit$\Thanks{work done while at Microsoft.},  Eduardo Gonzalez$^\spadesuit$ \\ {\bf Damien Jose$^\spadesuit$, Ahmed H. Awadallah$^\spadesuit$, Jianfeng Gao$^\spadesuit$}\\\\
        $^\spadesuit$Microsoft, $^\clubsuit$Google \\
        \normalsize$^\spadesuit$\texttt{\{shagup,submukhe,eduardogo,dajose,hassanam,jfgao\}@microsoft.com}\\ 
        \normalsize$^\clubsuit$\texttt{\{krishansubudhi\}@google.com}}

\begin{document}

\maketitle
\begin{abstract}
%% Abstract %%
Traditional multi-task learning (MTL) methods use dense networks that use the same set of shared weights across several different tasks. This often creates interference where two or more tasks compete to pull model parameters in different directions. In this work, we study whether sparsely activated Mixture-of-Experts (MoE) improve multi-task learning by specializing some weights for learning shared representations and using the others for learning task-specific information.
To this end, we devise task-aware gating functions to route examples from different tasks to specialized experts which share subsets of network weights conditioned on the task. This results in a sparsely activated multi-task model with a large number of parameters, but with the same computational cost as that of a dense model. We demonstrate such sparse networks to improve multi-task learning along three key dimensions: (i) transfer to low-resource tasks from related tasks in the training mixture; (ii) sample-efficient generalization to tasks not seen during training by making use of task-aware routing from seen related tasks; (iii) robustness to the addition of unrelated tasks by avoiding catastrophic forgetting of existing tasks.

%
%reuse {\em all} the weights for examples even from different tasks. In this work, we study whether sparsely activated networks improve multi-task learning for diverse NLU tasks which allow them to activate subsets of the neural network weights for different tasks as well as share weights across related tasks. 

%We demonstrate our sparse multi-task mixture-of-experts (\msparsemtl) model to significantly outperform traditional dense single-task and several dense and sparse multi-task models on key various aspects including low-resource task transfer, robustness to unrelated tasks, and generalization to unseen related tasks.

%"Towards this goal, we devise task-aware gating functions that utilize the task information from the input to guide it to experts from a shared pool."
\end{abstract}

\section{Introduction}
\label{sec:intro}
%% Introduction %%
%Large-scale pre-trained language models (PLMs) have shown state-of-the-art performance on several Natural Language Understanding (NLU) tasks. 
The traditional mechanism of using large-scale pre-trained language models PLMs~\citep{Devlin2019BERTPO, He2021DeBERTaDB} involve fine-tuning them for each task individually. This approach %is not only computationally expensive since it requires updating a large number of model parameters for each task, but also 
fails to benefit from interactions between tasks that could be related to each other. For instance, the task of predicting if one text entails or contradicts another can benefit from tasks that predict whether two texts are semantically similar or not. To address these limitations, Multi-Task Learning (MTL) methods like MT-DNN~\citep{Liu2019MultiTaskDN} and Muppet~\cite{Aghajanyan2021MuppetMM} instead train a single model jointly on a multi-task mixture consisting of multiple tasks. The typical mechanism is to facilitate transfer between the tasks by encoding the examples using a task-agnostic network shared between all the tasks, and then using task-specific layers on top to optimize individual task objectives. The dominant choice for the network is a Transformer-based PLM such as BERT~\cite{Devlin2019BERTPO}. However, such dense (fully-connected) task-agnostic networks have the limitation that they use all the weights of the network for every example, including those coming from very different tasks. This creates interference among different tasks, e.g., the tug-of-war phenomenon~\cite{Hadsell2020EmbracingCC} where two or more tasks pull the model parameters in different directions, thus impacting the multi-task learning performance. %However, different tasks can require surfacing different information from the underlying text, however, such dense models produce task-agnostic representations, wherein the same text in different tasks is represented similarly. 
%unless the task information is explicitly provided through a task prompt or task embedding -- revisit if we need to explain why we don't just use a task-embedding or task-prompt -- they will still be hard to scale up

A possible mechanism to alleviate this problem is to devise a task-aware network that can capture specialized information about individual tasks, as well as information that can be shared among multiple tasks. Mixture-of-Experts (MoE) framework~\citep{Shazeer2017OutrageouslyLN, Fedus2021SwitchTS,lepikhin2021gshard} provides a way to model this mechanism. Such architectures are designed to support conditional computation in which only certain weights of the network are activated per input as governed by a gating mechanism. This sparse design has an additional advantage of providing additional capacity in terms of model parameters while keeping overall computational cost constant.

The above sparse MoE models have been typically trained from scratch using language modeling objectives for tasks like neural machine translation; or fine-tuned on NLU tasks in a single-task setting. In contrast, in this work {\em we study multi-task adaptation (as opposed to pre-training from scratch) of sparse MoE models {on diverse NLU tasks} when judiciously initialized with the weights of a pre-trained language model}. Our motivation for using MoEs is that the sparsity and conditional computation within MoEs will help to alleviate inter-task interference by specializing some weights for learning shared representations and using the others for learning task-specific information.

Multi-task adaptation for sparse MoE models that have been traditionally used in single-task settings require rethinking the gating mechanism. Existing sparse models use a single task-agnostic shared gate that learns to route inputs from all the tasks, leading to interference wherein different tasks compete for the shared gate.

\textbf{Contributions}: We ({\em Contribution 1)} first address this limitation by devising a task-aware gating mechanism within sparse MoEs to route the input (tokens from different tasks) to specialized experts conditioned on the task to support MTL. 

Thereafter, ({\em Contribution 2.1)} we perform an extensive empirical study of the robustness of dense and sparse models to inter-task interference for multi-task learning on three key dimensions, (i) {\em transfer to low-resource tasks} from related tasks in the training mixture; (ii) {\em sample-efficient generalization to tasks not seen during training} from related seen tasks; (iii) {\em robustness to the addition of unrelated tasks} by avoiding catastrophic forgetting of existing tasks. 
%by making use of task-aware routing from seen related tasks
%
We ({\em Contribution 2.2)} empirically demonstrate sparse MoE models with task-aware gating and routing to be more robust multi-task learners than their non-MoE dense counterparts on the above dimensions.

\section{Sparse Mixture-of-Experts: Background}
\label{ssec:mmoe_model}
\begin{figure*}[ht!]
\includegraphics[width=0.8\textwidth]{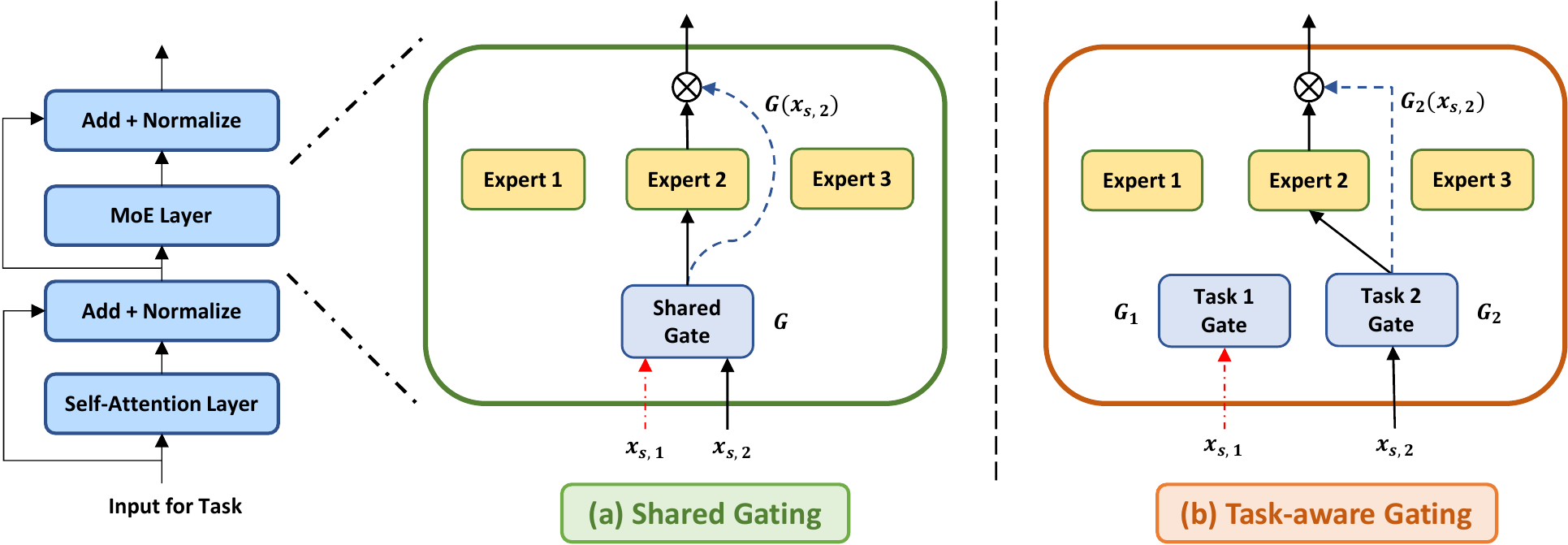}
\centering
\caption{Sparse MoE layer with $3$ Experts, $2$ Tasks, and $top$-$1$ expert routing with (a) Shared Gating, and (b) Task-aware Gating. $x_{s,1}$ and $x_{s,2}$ are tokens from Task $1$ and $2$ respectively. They share the same gate $G$ in sub-figure (a), and routed to respective task-specific gates in sub-figure (b). For simplicity, we only show the pathway for $x_{s,2}$ with a solid line, and show the gating behavior for $x_{s,1}$ with a dashed \textcolor{red}{red} line}
\label{fig:diagram}
\vspace*{-1.4em}
\end{figure*}

%% Sparse Multi-Task Learning %%
We adopt the popularly used Transformer architecture~\cite{vaswani2017attention} as the basic encoder %Transformers have been shown to obtain state-of-the-art performance in several natural language understanding tasks. 
consisting of $L$ repeated Transformer blocks, 
%Considering an input sequence of $n$ tokens $x=\{x_1, x_2, \cdots x_n\}$, the token embedding ${W}$ is added to the position ${PE}$ and segment ${SE}$ embeddings as $z_i(x_i) = W(x_i) + PE(i) + SE(i)$. The input to the network is given by  $\mathcal{H}^0=[z_1, z_2, \cdots z_{|x|}]$ and all the embeddings are learnable. 
%
%repeatedly computing hidden state representations from the output of the previous layer. %, where hidden states from the $l$th layer of the model is given by $\mathcal{H}_l = Transformer_l(\mathcal{H}_{l-1}),\ l \in [1, \cdots \mathcal{L}]$.
%
% {\small
% % \setlength\abovedisplayskip{0pt}
% % \setlength\belowdisplayskip{0pt}
% \begin{align}
% \mathcal{H}_l &= Transformer_l(\mathcal{H}_{l-1}), l \in [1, \cdots \mathcal{L}] \\
%     \mathcal{H}_l &= [{h}_{l,1}, {h}_{l,2}, \cdots, {h}_{l,|x|}], l \in [1, 2, \cdots \mathcal{L}]
% \end{align}
% }
%
%where, $h_{l,i}$ represents the hidden state representation of token $x_i$ from layer $l$. %
where each block consists of a self-attention sub-layer, a fully connected feed-forward network (FFN) and residual connections around the sub-layers followed by layer normalization.

The objective of sparse design of the above Transformer blocks is to support conditional computation and increase the parameter count while keeping the floating point operations (FLOPs) for each input example constant. Mixture-of-Experts (MoE) Transformer models~\citep{Shazeer2017OutrageouslyLN, Fedus2021SwitchTS,lepikhin2021gshard,Zuo2021TamingSA} achieve this by using $N$ feed-forward networks (FFN), namely ``experts" denoted as $E_{i=1}^N$, each with its own set of learnable weights. % that compute different representations of an input token $x$ based on context. 
In order to sparsify the network to keep the FLOPs constant, there is an additional gating network $G$ whose output is a sparse $N$-dimensional vector to route each token via a few of these experts. Note that, a sparse model with $N=1$ corresponding to only one FFN layer in each Transformer block collapses to the traditional dense model. %\sg{Check the comment here.}

Consider $x_s$ as the input token representation in the $s^{th}$ position to the MOE layer comprising of the $\{E\}_{i=1}^N$ expert FFNs. Also, consider $w^{in}_i$ and $w^{out}_i$ to be the input and output projection matrices for $i^{th}$ expert. Expert output  $E_i(x_s)$ is given by:

\begin{equation}
    E_i(x_s) = w^{out}_i \cdot GeLU (w^{in}_i \cdot x_s)
\end{equation}

\noindent Consider $G(x_s)$ to be output of the gating network. Output of the sparse MoE layer is given by:

\begin{equation}
\label{eq:moe}
    h(x_s)=\sum_i G(x_s)_i\ E_i(x_s) 
\end{equation}

\noindent where $G(x_s)_i$ denotes the probability of selecting expert $E_i$ for $x_s$. 
% \noindent where $G(x_s)_i$ the $i^{th}$ logit of the output of $G(x_s)$ denotes the probability of selecting expert $E_i$. 

%\sg{Include the concept of selecting only k experts, and k=1 leading to Switch Transformers. Top-k gating from the next section can be moved here as a fundamental part of Sparse MoEs.}

%In order to keep the number of FLOPs in the sparse Transformer to be the same as that of a dense one, the gating mechanism can be constrained to route each token to only one expert FFN, i.e. $\sum_i G_t(x_s)_i = 1$.

\section{Sparse Multi-task Learning with Mixture-of-Experts} %\msparsemtl} %({\small \msparsemtl)}}
\label{sec:mmoe}

We first highlight the shortcoming of existing sparse MoE models for multi-task learning and our architectural modifications to support the same along with an analysis of its impact on the model size and task scalability. We then present some details on the task formulation and optimization objectives to train sparse multi-task models.

\subsection{Task-aware Sparse Routing to Experts}
\label{ssec:mmoe_description}

The sparse MoE design outlined in the previous section does not consider the underlying task (Figure~\ref{fig:diagram}(a)). Given the same input from different tasks, the task-agnostic gating mechanism routes tokens to the same experts, thereby generating similar hidden-state representations. This is an issue during multi-task learning, where it is beneficial to learn task-specific contextualized representation of the input. To address this shortcoming, we modify the gating function to be task-aware, such that inputs from a given task are routed to specialized experts that also share weights across related tasks. 

Consider a set of ${T}$ diverse tasks in the multi-task mixture and $x_{s,t}$ to be the token representation in the $s^{th}$ position of the input sequence from task $t \in T$, where each task is equipped with its own loss function. 
% 
%{\noindent \bf Task-aware gating network:}
Consider trainable weight matrices $W_{g,t} \in \mathcal{R}^{N \times H}$ corresponding to each task $t \in T$ where, $N$ is the number of experts and $H$ is the hidden state dimension. To incorporate task information in the gating mechanism, we multiply the input $x_{s,t}$ with the task-specific weight matrix $W_{g,t}$ to obtain the routing logits:

\begin{equation}
 {l}_t(x_{s,t}) = x_{s,t} \cdot W_{g,t}^\mathsf{T}
\end{equation}

We can further normalize them via a softmax distribution over the $N$ experts in each MoE layer to obtain the corresponding routing probabilities. The gate-value for the $i^{th}$ expert is given by:

\begin{equation}
    G_t(x_{s,t})_i = \frac{e^{{l}_t(x_{s,t})_i}}{\sum_{j=1}^N e^{{l}_t(x_{s,t})_j}}
\end{equation}

We can now select the $top$-$k$ gate values for routing the token. In order to keep the number of FLOPs in the sparse Transformer to be the same as that of a dense one, the gating mechanism is constrained to route each token to only the $top$-$1$ expert FFN selected as:

\begin{equation}
 {g}^*_t(x_{s,t}) = max_i\ G_t(x_{s,t})_i   
\end{equation}

The output of the sparse MOE layer in Equation~\ref{eq:moe} can be modified with the task-specific gating function by linearly combining the selected $top$-$1$ expert's ($E^*$) computation on $x_{s,t}$ and the probability of selecting the expert as:

\begin{equation}
    h(x_{s,t}) = {g}^*_t(x_{s,t})\ E^*(x_{s,t}) 
\end{equation}

where $h$ denotes the task-specific representation of input $x_{s,t}$. 

In the above formulation, the task-specific gating function $G_t$ learns to route tokens from the input to specialized experts. Note that the experts themselves do not have explicit relationship with the task and are only dependent on input context so as to encourage information sharing among all experts. The expert selection is implicitly conditioned on the task id $t$ (provided with the input) via task-aware gating function $G_t$. We refer our framework as \textbf{\msparsemtl}, short for \underline{M}ulti-\underline{T}ask \underline{T}ask-\underline{a}ware \underline{G}ating (Figure~\ref{fig:diagram}(b)).
%This rewards the selection of similar experts for tokens from tasks that are related to each other, and penalizes sharing the experts otherwise. 

\subsection{Analysis of Sparsity and Task-scalability} %The advantage of the above design in addition to flexible weight sharing and task-specificity is the introduction of very few task-specific parameters in the sparse model.
\label{ssec:mmoe_analysis}

We introduce the feed-forward networks (FFN) as experts in every layer of the Transformer. Consider $N$ experts, $L$ layers and $P_f$ to be the number of parameters in each FFN expert. The number of expert parameters in the model is $L \times N \times P_f$. Since the experts are shared among all tasks, increasing the number of tasks does not impact expert parameters.

On the other hand, the gating network is task-aware which increases the number of parameters with more tasks. Considering $H$ to be the hidden state dimension and $T$ to be the number of tasks, the number of gating parameters is $L \times N \times H \times T$.

Since the hidden state dimension and number of tasks are much less than the number of FFN parameters (i.e., $H \times T \ll P_f$) in most practical settings, increasing tasks contribute very less parameters as compared to the parameters already contained in the standard feed-forward Transformer networks.  

Consider the following as an illustration. Consider a $6$-layer Transformer with $384$ hidden dimension and {\bf $22M$ encoder parameters} corresponding to a standard dense Transformer. Consider $4$ experts and $8$ tasks for MTL, where we introduce these experts in each Transformer layer. \msparsemtl contains only {\bf $74K$ gating parameters} in the task-specific gating networks for expert selection as compared to {\bf $21M$ expert parameters}. In total, the sparse \msparsemtl model doubles the number of parameters as compared to the dense model although incurring the same number of FLOPs with $top$-$1$ expert selection. This capacity coupled with task-awareness improves model performance in MTL as demonstrated in experiments. 

\subsection{Multi-task Training}
\label{ssec:mmoe_training}
%\sg{There's currently an abrupt jump here, we need to connect this subsection with the previous with something like .. in order to train this sparse architecture on a mixture of tasks, we add task-specific ..}
We now outline multi-task objectives and protocol for training the \msparsemtl model.

\noindent {\bf Task objectives:} For a classification task $t$, we use a task-specific projection layer on top of the MTL encoder to obtain the class probability distribution for the contextualized representation of an input example $x_t$\footnote{For inputs with sequence pairs ($x^1$, $x^2$), we consider $x= x^1 \oplus x^2$, with $\oplus$ representing concatenation operation.} from task $t$ as:
%or an input pair of examples $x_t= x_{t,1} \oplus x_{t,2}$, with $\oplus$ representing concatenation as:

\begin{equation}
    P(c|x_t) = Softmax(U_t \cdot h(x_t))
\end{equation}

where, $U_t \in \mathbb{R}^{C_t \times d}$ is the task-specific parameter matrix with $C_t$ representing the number of classes and $d$ as the hidden state dimension.

For a regression task $t$ (e.g., textual similarity), we obtain the output score for the contextualized representation of the input $x_t$ as:
%pair of examples as $x_t= x_{t,1} \oplus x_{t,2}$, with $\oplus$ representing concatenation:

\begin{equation}
    S(x_t) = V_t \cdot h(x_t)
\end{equation}

where, $V_t \in \mathbb{R}^{1 \times d}$ is the task-specific parameter matrix and $S(x_t) \in \mathbb{R}(-\infty, \infty)$.

For classification tasks, we use cross-entropy loss, where we train the network to minimize the following objective in the MTL setup:

\begin{equation}
\label{eq:classification}
    - \sum_{t \in \mathbb{T}} \sum_{x_t \in X_t} \sum_{c \in C_t}  \mathbbm{1}(x_t,c)\ log\  P(c|x_t)
\end{equation}

where, $X_t$ is the set of examples from task $t$, $\mathbbm{1}(x,c)$ is the binary indicator which is $1$ if $c$ is the correct class label for $x$ and $0$ otherwise. 

For regression tasks, we use mean-squared error loss, where we train the network to minimize the following objective in the MTL setup:

\begin{equation}
\label{eq:regression}
     \sum_{t \in \mathbb{T}} \sum_{\langle x_t, y_t \rangle \in \langle X_t, Y_t \rangle} (y_t- S(x_t))^2
\end{equation}

where, $\langle X_t, Y_t \rangle$ is the set of examples from task $t$ with corresponding ground-truth scores. 

\noindent {\bf Joint optimization:} We jointly optimize Equations~\ref{eq:classification} and~\ref{eq:regression} to train the entire model including the gating network by back-propagation, where the gradients back-propagate through the gating network to the inputs.

\noindent{\bf Loss scaling:} In the MTL setup, the number of classes per task can vary. To ensure stability in the training, we leverage loss scaling to normalize the task-specific loss function in Equation~\ref{eq:classification} with respect to the number of classes in the task $t$ as $\big( \sum_{c \in C_t}  \mathbbm{1}(x_t,c)\ log\  P(c|x_t) \big) / log(|C_t|)$, where $|.|$ denotes the cardinality of the set of classes.

\noindent{\bf Batching and sampling:} The MTL training process optimizes several objectives which are often at loggerheads with each other. %We evaluate several batching policies that have been found to be effective in prior works. For instance, MT-DNN~\cite{Liu2019MultiTaskDN} employs {\em homogeneous batching}, such that the scheduler first selects a task and then a set of samples from the task to optimize the task-specific objective, updates the model before moving on to the next task. 
Recent work~\cite{aghajanyan-etal-2021-muppet} demonstrates {\em heterogeneous batching} to work better for MTL, where batches from different tasks are sampled to construct a super-batch, which is then used for jointly optimizing corresponding task-objectives. We follow similar principles along with employing a natural sampling of tasks, wherein we sample batches from tasks in proportion to their dataset sizes to reflect the complexity of the corresponding tasks.

\section{Experimental Setup}
\label{sec:exp_setup}
%% Experimental Setup%%
% \subsection{Evaluation Criteria}
% \label{ssec:eval_criteria}
% \sg{Robustness, Transfer to low-resource tasks, Domain Adaptation, Task scaling, Encoder scaling}
% We identify several characteristics desirable in a good Multi-Task Learning (MTL) system, and use them as our evaluation criteria. In particular, a good MTL system:
% \begin{enumerate}
%     \item \textbf{Efficient Transfer}: Should be able to benefit similar tasks in the training mixture, especially the low-resource ones.
%     \item \textbf{Generalization}: Should be able to generalize and improve the performance of tasks not seen during the MTL training that have similarities with tasks in the training mixture.
%     \item \textbf{Robustness}: Should be robust, in that addition of unrelated tasks shouldn't hurt the performance of tasks already in the training mixture.
%     \item \textbf{Encoder size scaling}: Should show gains across encoder sizes.
%     \item \textbf{Task scaling}: Should be able to show gains even if the number of tasks in the mixture is scaled up.
% \end{enumerate}

% \subsection{Datasets}
% \label{ssec:exp_datasets}
% GLUE
% SuperGLUE
% IMDB, Yelp Polarity
% SciTail
% We provide additional details about the datasets in ~\ref{app:datasets}

\subsection{Datasets}
\label{ssec:exp_datasets}
We use $8$ diverse NLU datasets from the GLUE benchmark~\cite{Wang2018GLUEAM} for MTL training consisting of single-text classification tasks such as COLA and SST-2; paired-text classification tasks such as RTE, MRPC, QNLI, QQP, and MNLI; and paired-text regression tasks such as \mbox{STS-B}. These evaluate various NLU capabilities such as sentiment classification in SST-2; textual entailment in RTE, QNLI, and MNLI; paraphrase detection in MRPC and QQP; text similarity in STS-B; and text acceptability in CoLA. 
There are varying number of examples per dataset ranging from $2.5K$ examples in the smallest one (RTE) to $393K$ examples in the largest one (MNLI). This allows us to study the efficacy of MTL models in terms of transfer to low-resource tasks. The task mixture also consists of tasks like COLA and SST-2 that have low similarity with the rest, enabling us to study the robustness of MTL models in the presence of unrelated tasks. We provide more details about these datasets and their sizes in Appendix \ref{app:datasets} and Table~\ref{tab:all_dataset_stats}.
% \input{tables/glue_dataset_stats}.

%We also use these datasets to construct our multitask mixture for MTL training. This multitask mixture contains related datasets of varying sizes, for instance MRPC and QQP are both paraphrasing datasets with $3.7$k and $364$k training examples respectively, and similarly, RTE and MNLI are both textual entailment datasets with $2.5$k and $393$k training examples respectively. This variation in the sizes of related datasets allows us to study the low-resource task transfer efficacy of the MTL models. Furthermore, the varying degree of similarity between the tasks, such that CoLA and SST2 don't have much similarity with the rest, makes it a good multitask mixture to study the robustness of MTL models as well. We provide more details about these datasets and their sizes in Appendix \ref{app:datasets} and Table~\ref{tab:dataset_stats}
% \input{tables/glue_dataset_stats}
%%%%%%%%%%%%%%%%%%%%%%%%%%%%%%%%%%%%%%%%%
%%%%%%%%%%%%%%%%%%%%%%%%%%%%%%%%%%%%%%%%%
\subsection{Models for Comparison}
\label{ssec:exp_baselines}
%%%
We consider several models that are all FLOPs matched per token for comparison as follows.

\noindent{\textbf{(a) \singletask}}: This baseline trains a dense model directly on individual end-tasks without MTL. Since there is no interaction across tasks, this baseline helps us evaluate the impact of MTL.

%%%
\noindent{\textbf{(b) \densemtl}}: This baseline is created by training a dense MTL model. %This baseline helps us study the impact of sparsity induced MTL training versus that of a dense one.
Note that this baseline is similar in flavor to the multi-task learning methods like MT-DNN~\cite{Liu2019MultiTaskDN} and Muppet~\cite{aghajanyan-etal-2021-muppet}.

%%%
\noindent{\textbf{(c) \sparsemtl}}: This is a sparse MTL Mixture-of-Experts model using a single shared gate for all tasks as depicted in Figure~\ref{fig:diagram}(a). Note that \sparsemtl differs with \msparsemtl only in its usage of a single task-agnostic shared gate, helping us evaluate the impact of task-aware gating.

\noindent{\textbf{(d) \msparsemtl}}: This is the sparse MTL Mixture-of-Experts model outlined in Section~\ref{ssec:mmoe_description} (depicted in Figure~\ref{fig:diagram}(b)) that uses task-aware gating.

All the models have similar FLOPs per token and all the MTL models are trained using the procedure outlined in Section~\ref{ssec:mmoe_training}. We use $top$-$1$ expert routing for both sparse MTL models.

% We follow \msparsemtl's model selection methodology, tuning procedures, and hyper-parameter ranges for all the baseline models.
%%%%%%%%%%%%%%%%%%%%%%%%%%%%%%%%%%%%%%%%%
% \subsection{Task formulations}
% \label{ssec:task_formulations}
% We add a [CLS] token at the beginning of the input for all the datasets. For paired-text classification and regression tasks, we concatenate the 2 sequences separated by a [SEP] separator token. We use the [CLS] token's output representation from the final layer of the encoder as the representation of the input, which is then passed to the task-specific classification and regression layers (collectively referred to as \emph{task-heads} henceforth) to optimize for the corresponding task-objective.
%%%%%%%%%%%%%%%%%%%%%%%%%%%%%%%%%%%%%%%%%
\subsection{Model Initialization and Setup}
\label{ssec:exp_sparse_encoders}
%\sg{TODO: This section might require some cleanup, or/and section name change. The aim is to cover the following aspects: 1) The Xsmall encoder 2) the relation between dense and their sparse encoder counterparts 3) how we initialize the Sparse model using the weights of a dense pre-trained model}

% To reap the benefits of transfer learning, we initialize our dense and sparse models using the weights of dense pre-trained language models.

\textbf{Dense models}: As in prior multi-task learning works~\citep{Liu2019MultiTaskDN}, we initialize the dense model using weights from pre-trained language models. 
%
%To reap the benefits of transfer learning, we initialize our dense models using the weights from dense pre-trained language models. 
In addition to using \bertbase ($12$ layers, $768$ hidden size, $110M$ params) and \bertlarge ($24$ layers, $1024$ hidden size, $345M$ params) pre-trained models, we also consider \encodername~\cite{Wang2021MiniLMv2MS} ($6$ layers, $384$ hidden size, $22M$ params) distilled from $\mbox{BERT}_{Large}$ as its compressed variant. Unless otherwise stated, we use \encodername as our default encoder to carry out an extensive study with limited compute resources.

%create a smaller transformer model containing $22$M parameters to carry out an extensive study with limited compute resources. We call this the \textbf{\encodername} encoder to contrast it with the $110$M parameter $\mbox{BERT}_{Base}$ and $335$M parameter $\mbox{BERT}_{Large}$ models. It consists of a stack of $6$ identical transformer blocks, $12$ attention heads, $384$ hidden size, $1536$ feed-forward layer size, and is initialized using the weights of the $22$M model checkpoint released by MiniLMv2~\cite{Wang2021MiniLMv2MS}, distilled from a $\mbox{BERT}_{Large}$ teacher model. Unless otherwise stated, \encodername is our default encoder in the experiments.

\noindent \textbf{Sparse models}: For a fair comparison with the dense models, we create FLOPs matched sparse models, and initialize them using the weights of dense pre-trained language models. To this end, we replace the feed-forward layers (FFNs) in each transformer layer of the dense model with a MoE layer containing $N$ experts and $T$ gates ($T=1$ for \sparsemtl; $T=\text{num. of tasks}$ for \msparsemtl). This results in as many MoE layers as the number of Transformer layers of the corresponding dense pre-trained language model used for initialization. 
To initialize the FFN weights of experts in any MoE layer, we simply make $N$ copies of the FFN weights of the corresponding layer from the dense pre-trained language model\footnote{Experiments with initializing expert weights differently by adding a small random noise did not show improvements.}.

\begin{table*}[ht]
\centering
\resizebox{\linewidth}{!}{%
\begin{tabular}{ccccccccccc} 
\toprule
Model & \begin{tabular}[c]{@{}c@{}}RTE\\\small{(2.5k)}\end{tabular} & \begin{tabular}[c]{@{}c@{}}MRPC\\\small{(3.7k)}\end{tabular} & \begin{tabular}[c]{@{}c@{}}STS-B\\\small{(5.7k)}\end{tabular} & \begin{tabular}[c]{@{}c@{}}CoLA\\\small{(8.5k)}\end{tabular} & \begin{tabular}[c]{@{}c@{}}SST-2\\\small{(67.3k)}\end{tabular} & \begin{tabular}[c]{@{}c@{}}QNLI\\\small{(105k)}\end{tabular} & \begin{tabular}[c]{@{}c@{}}QQP\\\small{(364k)}\end{tabular} & \begin{tabular}[c]{@{}c@{}}MNLI\\\small{(393k)}\end{tabular} & \begin{tabular}[c]{@{}c@{}}Small Tasks\\\small{(Avg.)}\end{tabular} & \begin{tabular}[c]{@{}c@{}}All Tasks\\\small{(Avg.)}\end{tabular} \\ 
\midrule\midrule
\singletask & 70.7 & 88.7 & 88.9 & \uline{41.8} & \textbf{92.4} & \textbf{90.4} & 90.6 & \textbf{83.9} & 72.5 & 80.9 \\
\midrule
\densemtl & 77.9 & 89.0 & \uline{90.5} & \textbf{42.1} & 92.0 & \uline{90.3} & \uline{90.8} & \uline{83.8} & 74.9 & 82.1 \\ 
\sparsemtl & \uline{78.9} & \uline{90.0} & \uline{90.5} & 40.7 & 92.0 & \uline{90.3} & \textbf{90.9} & 83.6 & \uline{75.0} & \uline{82.1} \\
\midrule
\textbf{\msparsemtl} & \textbf{81.1} & \textbf{90.7} & \textbf{90.6} & 41.1 & \uline{92.1} & 90.2 & \uline{90.8} & 83.6 & \textbf{75.9} & \textbf{82.5} \\
\bottomrule
\end{tabular}
}
\vspace{-0.5em}
\caption{Comparison of dense and sparse models on GLUE. Best task numbers are \textbf{boldfaced}, and second-best \uline{underlined}. Sparse MoE with task-specific gating (\msparsemtl) outperforms \singletask and FLOPs matched dense and sparse MTL models with significant improvements for low-resource tasks. All models use MiniLM encoder.}
\label{tab:glue}
\vspace{-1em}
\end{table*}

\subsection{Implementation Details}
\label{ssec:exp_imp_details}
We use standard wordpiece tokenization ($30K$ vocabulary) and segmentation for the input sequences. We use $N=4$ experts in all layers for our experiments\footnote{We provide results with varying \#experts in Appendix.}, giving us sparse models with $44M$, $280M$, and $940M$ parameters that are FLOPs matched to \encodername, \bertbase, and \bertlarge encoders, respectively. We initialize all gating weights using a normal distribution with $0$ mean and $0.001$ standard deviation. Similarly, we initialize task-specific parameter matrices $\mathbb{U}_t,\mathbb{V}_t$ using a normal distribution with $0$ mean and $0.02$ standard deviation. We initialize all layer normalization weights with $1$, bias weights with $0$, and use a dropout of $0.1$.

We use Adam Optimizer~\cite{Kingma2015AdamAM} with a linear learning rate decay schedule and warm-up. We use mixed-precision training, clip the norms of gradients to $1$, and use $4$ Nvidia V100 GPUs for distributed training. We utilize PyTorch and HuggingFace Transformers~\cite{Wolf2019HuggingFacesTS} for our implementation\footnote{Our code and model checkpoints will be made public.}.
%We provide all hyper-parameter values in Appendix ~\ref{app:tuning}
%\footnote{https://www.nvidia.com/en-us/data-center/v100/}
%PyMarlin\footnote{https://github.com/microsoft/PyMarlin}
%~\cite{Paszke2019PyTorchAI}
% \sg{<<For MTL Training: Choice of Batching and Sampling, Choice of loss function}
%%%%%%%%%%%%%%%%%%%%%%%%%%%%%%%%%%%%%%
%%%%%%%%%%%%%%%%%%%%%%%%%%%%%%%%%%%%%%
\subsection{Evaluation}
\label{ssec:exp_eval}
% \label{ssec:exp_procedure}
\textbf{MTL Training protocol}: We follow a two-stage training protocol for MTL models. We first train the dense or sparse model (initialized from a pre-trained language model as outlined in Section~\ref{ssec:exp_sparse_encoders}) on a multitask mixture such as the GLUE dataset following the MTL training procedure (as outlined in Section~\ref{ssec:mmoe_training}) for a fixed number of steps, which gives us the corresponding MTL model. We then further fine-tune the MTL model on individual target datasets. This additional fine-tuning step has been shown to be beneficial for the model performance \cite{Liu2019MultiTaskDN}. Note that we use the same training protocol for all the MTL models.

\noindent{\textbf{Metrics}}: We use the standard train and dev splits for all GLUE datasets for training and evaluation. For the MTL models, we report the numbers obtained from the fine-tuning stage. We use Spearman correlation as our evaluation metric for STS-B, Matthews correlation coefficient (MCC) for COLA, and accuracy for the rest. For MNLI, we report the average accuracy on the matched (in-domain) and mismatched (cross-domain) splits. We additionally report two aggregate statistics: \emph{All Tasks}, and \emph{Small Tasks}, capturing the average performance on all tasks and just the small tasks respectively. We define Small Tasks as the tasks with $\leq10k$ examples, which for GLUE includes RTE, MRPC, STS-B, and COLA.
We provide more experimental details, including hyper-parameter tuning and values in Appendix \ref{app:tuning}.
%We choose task/random switches and head>>

\section{Robustness Analysis}
\label{sec:results}
%% Results %%
%%%% Low-resource transfer section %%%%
% \sg{Consider adding an overarching experimentation plan here -- what capabilities we study and why}
We perform an extensive empirical study of the robustness of sparse and dense MTL models along key dimensions with the following desiderata:

\noindent \circled{1} \textbf{Transfer to low-resource tasks}: A robust model should be able to alleviate task interference in the training mixture and improve performance on the low-resource tasks through transfer from other related tasks.

    % \item \textbf{Transfer to low-resource tasks}: A robust model should be able to alleviate task interference in the presence of high-resource tasks in the training mixture and improve performance on the low-resource ones.
    
\noindent \circled{2} \textbf{Sample-efficient generalization to unseen related tasks}: A robust model should be able to retain information from individual tasks in its training mix, and generalize in a sample-efficient manner to new related tasks that are not seen during training.

\noindent \circled{3} \textbf{Robustness to the addition of unrelated tasks}: A robust model should be better at weathering the interference introduced by the addition of unrelated tasks in its training mixture, and avoid catastrophic forgetting of existing tasks.

\subsection{Low-resource Task Transfer}
\label{ssec:direct_ft_results}

We first evaluate the ability of MTL models to leverage task-level similarities in the multitask mixture to improve performance on low-resource tasks. To this end, we train and evaluate all models on GLUE. Table~\ref{tab:glue} shows that all MTL models obtain improvements on low-resource tasks over \singletask baseline, while maintaining similar performance on relatively high-resource tasks. This demonstrates the benefit of multi-task learning in utilizing inherent similarities between tasks. Furthermore, we observe that both the sparse MoE models (\sparsemtl and \msparsemtl) outperform the non-MoE dense one (\densemtl), demonstrating the benefit of inducing sparsity for MTL. Finally, we observe the sparse MoE model with task-aware gating (\msparsemtl) to outperform all baselines, including single-gate sparse MoE (\sparsemtl), demonstrating improved ability to mitigate interference between tasks during multi-task learning.
%high- and low-resource

%%%%%%%%%%%%%%%%%%%%%%%%%%%%%%%%%%%%%%
%%%%%%%%%%%%%%%%%%%%%%%%%%%%%%%%%%%%%%
%%%% Generalization section %%%%

%% Table 2: OOD generalization %%
% \begin{table*}
% \centering
% \begin{tabular}{c|c|cc} 
% \hline
% \multicolumn{2}{c|}{} & 10\% & 100\% \\ 
% \hline
% \multicolumn{2}{c|}{Bert-base} & 90.5 & 94.3 \\
% \multicolumn{2}{c|}{MT-DNN} & 91.1 & 95.7 \\
% \multicolumn{2}{c|}{MOE} & 91.3 & 95.0 \\ 
% \hline
% \hline
% \multicolumn{2}{c|}{MMOE-MTL} &  &  \\ 
% \cline{1-2}
% Head & Gate &  &  \\ 
% \hline
% Random & Random & 91.0 & 95.9 \\
% Random & RTE & 91.6 & 96.0 \\
% RTE & RTE & 90.1 & 94.3 \\
% \hline
% \end{tabular}
% \caption{OOD Generalization on SciTail}
% \label{tab:ood}
% \end{table*}

%TODO

\begin{table}
\centering
\small
\resizebox{0.8\columnwidth}{!}{%
\begin{tabular}{ccccc} 
\toprule
\multirow{2}{*}{Model} & \multicolumn{2}{c}{SciTail} & \multicolumn{2}{c}{IMDB} \\ 
\cmidrule{2-3}
\cmidrule{4-5}
 & \begin{tabular}[c]{@{}c@{}}\small{1\%}\\\scriptsize{(235)}\end{tabular} & \begin{tabular}[c]{@{}c@{}}\small{10\%}\\\scriptsize{(2.4k)}\end{tabular} & \begin{tabular}[c]{@{}c@{}}\small{1\%}\\\scriptsize{(250)}\end{tabular} & \begin{tabular}[c]{@{}c@{}}\small{10\%}\\\scriptsize{(2.5k)}\end{tabular} \\ 
\midrule\midrule
\singletask & 81.9 & 90.6 & 86.1 & 90.6 \\
\midrule
\densemtl & 86.8 & \textbf{93.3} & \uline{89.8} & \textbf{91.2} \\ 
\sparsemtl & \uline{89.3} & \uline{92.9} & \uline{89.8} & \uline{91.1} \\
\midrule
\textbf{\msparsemtl} & \textbf{90.0} & \uline{92.9} & \textbf{90.3} & \textbf{91.2} \\
\bottomrule
\end{tabular}
}
\vspace{-0.5em}
\caption{Generalization performance on low-resource unseen related tasks. \msparsemtl delivers large gains over \singletask, and outperforms other MTL models in extremely low-resource settings demonstrating superior sample-efficiency. All models use MiniLM encoder.}
\label{tab:ood}
\vspace{-1em}
\end{table}
%Sparse MoE with task-specific gating

\subsection{Sample-efficient Generalization to Unseen Related Tasks}
\label{ssec:ood_results}
Section~\ref{ssec:direct_ft_results} demonstrates the benefit of sparse models on improving the MTL model performance on low-resource tasks. In this experiment, we want to evaluate their ability to generalize to related tasks that were not encountered during MTL training in a sample-efficient manner. %However, it is expensive to re-train the MTL model for every new incoming task. Thus, it is essential that MTL models learn representations that generalize well to related tasks that are not a part of its original training mix. 

To study this generalization ability, we leverage SciTail and IMDB as the unseen tasks for the GLUE-trained MTL models. Note that these tasks have some similarity to a subset of the GLUE tasks. For instance, SciTail is an NLI dataset with similarities to RTE, QNLI, and MNLI in GLUE; whereas IMDB is a sentiment classification dataset with similarities only to SST-2. This variation in similarity helps us study the degree of transferability from the multi-task training mixture to the new unseen tasks. We simulate low-resource settings by creating $1\%$ and $10\%$ samples from these datasets to study sample-efficiency, yielding datasets with roughly $250$ and $2.5$k examples respectively. We use accuracy as the metric for both datasets. We provide more details about these datasets and their task formulation in Appendix~\ref{app:datasets} and Table~\ref{tab:all_dataset_stats}.

We only fine-tune the GLUE-trained MTL models on these datasets, and compare against corresponding \singletask baselines. For fine-tuning \msparsemtl, we exploit task-specific gates, and re-use the gate corresponding to SST-2 for IMDB, and the gate corresponding to MNLI for SciTail due to their task-level similarities.

Table~\ref{tab:ood} shows that all MTL models obtain improvements over the \singletask baselines, demonstrating generalization ability of the MTL models. Furthermore, we observe that \msparsemtl outperforms all baselines on extremely low-resource settings on unseen datasets {\bf demonstrating superior sample-efficiency} of sparse models. \msparsemtl shows improvements even on IMDB which has only one related dataset in GLUE demonstrating improved task transfer from related tasks. \emph{We attribute these capabilities to the re-use of \msparsemtl's task-specific gates and routing that help it to better transfer information from related tasks in a sample-efficient manner}. We further found re-using unrelated task gates and randomly initializing the gates to perform significantly worse (results in Appendix~\ref{ssec:diff_gates}).%, and surface it through the related switch re-usage for the unseen tasks.
%%%%%%%%%%%%%%%%%%%%%%%%%%%%%%%%%%%%%%
%%%%%%%%%%%%%%%%%%%%%%%%%%%%%%%%%%%%%%
%%%% Robustness section %%%%
\subsection{Robustness to Unrelated Tasks}
\label{ssec:robustness_results}
Section~\ref{ssec:ood_results} demonstrates the improved performance of sparse MTL models to transfer information from even a single task of its kind (referred to as \emph{singleton tasks} henceforth) in the multi-task mixture. %This encourages adding many diverse singleton tasks to the multitask mixture. 
In this section, we further evaluate the robustness of MTL models on adding several diverse singleton tasks. Specifically, we evaluate if the singleton task addition has an adversarial affect on the performance of existing tasks in the multi-task mixture due to catastrophic forgetting.

To study this, we remove CoLA and SST-2 singleton datasets from the GLUE multi-task mixture, and refer to this new clean multi-task mixture as \sixtasks (short for Clean-GLUE). We evaluate the robustness by training all MTL models on both GLUE and \sixtasks, and comparing their performance on the common tasks: RTE, MRPC, STS-B, QNLI, QQP, and MNLI. We report the average performance on the common \emph{Small Tasks} and \emph{All Tasks} in Table~\ref{tab:robustness}, and provide the corresponding task-level results in Table~\ref{tab:full_robustness} of Appendix~\ref{app:full_robustness}.

We observe performance of dense MTL model (\densemtl) to decrease from \sixtasks to GLUE, demonstrating its lack of robustness to unrelated datasets in the multi-task mixture. Both sparse MTL models show better robustness because of their capability to specialize experts for unrelated tasks. \msparsemtl performs the best, further demonstrating the usefulness of combining expert specialization in sparse MoE with task-specific routing.

This result, combined with the findings in Section~\ref{ssec:ood_results} demonstrate that \emph{\msparsemtl is not only better at transfer from singleton tasks, but is also more robust to their presence in the multi-task mixture}. This motivates scaling \msparsemtl to a large number of diverse tasks as demonstrated in Section~\ref{ssec:task_scaling_results}.

%TODO
% \begin{table}
% \centering
% \resizebox{\columnwidth}{!}{%
% \begin{tabular}{c|cc||cc} 
% \hline
% Model & \multicolumn{2}{c||}{Small Tasks} & \multicolumn{2}{c}{All Tasks} \\ 
% \cline{2-5}
%  & \sixtasks & \eighttasks & \sixtasks & \eighttasks \\ 
% \hline\hline
% \densemtl & 86.27 & 85.80\color{red}{$\downarrow$} & 87.18 & 87.05\color{red}{$\downarrow$} \\
% \hline\hline
% \sparsemtl & 86.27 & 86.47\color{ForestGreen}{$\uparrow$} & 87.22 & 87.37\color{ForestGreen}{$\uparrow$} \\
% \hline\hline
% \textbf{\msparsemtl} & 86.50 & 87.47\color{ForestGreen}{$\uparrow$} & 87.32 & 87.83\color{ForestGreen}{$\uparrow$} \\
% \hline
% \end{tabular}
% }
% \caption{Robustness results for the \encodername encoder. Sparse MTL models show robustness to the presence of unrelated tasks, with \msparsemtl being the most robust.}
% \label{tab:robustness}
% \end{table}

\begin{table}
\centering
\small
\begin{tabular}{ccc}
\toprule
\multicolumn{1}{c}{Dataset}       & \multicolumn{1}{c}{Small Tasks}                                             & All Tasks \\
\midrule\midrule
\multicolumn{3}{c}{$\densemtl$}                                                                                               \\ \midrule
\multicolumn{1}{c}{$\sixtasks$}   & \multicolumn{1}{c}{86.27}                                                   & 87.18     \\
\multicolumn{1}{c}{$\eighttasks$} & \multicolumn{1}{c}{85.80 \small{\color{red}{(-0.47)}}} & 87.05     \\
\midrule\midrule
\multicolumn{3}{c}{$\sparsemtl$}                                                                                              \\ \midrule
\multicolumn{1}{c}{$\sixtasks$}   & \multicolumn{1}{c}{86.27}                                                   & 87.22     \\
\multicolumn{1}{c}{$\eighttasks$} & \multicolumn{1}{c}{86.47 \small{\color{ForestGreen}{(+0.20)}}} & 87.37     \\
\midrule\midrule
\multicolumn{3}{c}{\textbf{$\msparsemtl$}}                                                                                             \\ \midrule
\multicolumn{1}{c}{$\sixtasks$}   & \multicolumn{1}{c}{86.50}                                                   & 87.32     \\
\multicolumn{1}{c}{$\eighttasks$} & \multicolumn{1}{c}{87.47 \small{\textbf{\color{ForestGreen}{(+0.97)}}}} & 87.83 \\ \bottomrule
\end{tabular}
\vspace{-0.5em}
\caption{Model performance on GLUE (containing several diverse tasks) and \sixtasks (as a subset of GLUE containing only related tasks) evaluated on the common tasks in both. Sparse MTL models demonstrate robustness in the presence of unrelated tasks in GLUE, with \msparsemtl with task-specific routing being the most robust. All models use MiniLM encoder.}
\label{tab:robustness}
\vspace{-1.2em}
\end{table}
% 
%%%%%%%%%%%%%%%%%%%%%%%%%%%%%%%%%%%%%%
%%%%%%%%%%%%%%%%%%%%%%%%%%%%%%%%%%%%%%

\section{Scaling Analysis}
\label{sec:scaling}
%%%% Encoder size scaling section %%%%
\subsection{Encoder Size Scaling}
\label{ssec:encoder_size_results}
We study the sensitivity of the \msparsemtl model performance with change in the encoder size. To this end, we train \msparsemtl using \encodername, \bertbase and \bertlarge encoders of varying number of parameters. From Table~\ref{tab:enc_size}, we observe that \msparsemtl significantly outperforms single-task baselines across different encoder sizes. 

We also compare against the multi-task MT-DNN model from \citealp{Liu2019MultiTaskDN}, which is similar in flavor to our \densemtl model. Our sparse MTL MoE model \msparsemtl shows impressive gains over the dense MT-DNN model\footnote{MT-DNN only provides numbers for \bertlarge.}, especially on low-resource tasks. We provide task-level results for comparison in Table~\ref{tab:full_enc_size} of Appendix~\ref{app:full_encoder_size}.
%% Table: Showing performance gains all over the board%%
% \begin{table*}
% \centering
% \resizebox{\linewidth}{!}{%
% \begin{tabular}{c|c|c||c|c||c|c} 
% \hline
% \multirow{2}{*}{\begin{tabular}[c]{@{}c@{}}Model\\\end{tabular}} & \multicolumn{2}{c||}{\encodername} & \multicolumn{2}{c||}{$\mbox{BERT}_{Base}$} & \multicolumn{2}{c}{$\mbox{BERT}_{Large}$} \\ 
% \cline{2-7}
%  & \small{Small tasks} & \small{All tasks} & \small{Small tasks} & \small{All tasks} & \small{Small tasks} & \small{All tasks} \\ 
% \hline
% \singletask & 72.53 & 80.93 & 76.53 & 83.34 & 78.85 & 84.93 \\
% \textbf{\msparsemtl} & \textbf{75.88} & \textbf{82.53} & \textbf{80.73} & \textbf{85.45} & \textbf{82.73} & \textbf{86.94} \\
% \hline
% \end{tabular}
% }
% \caption{Impact of encoder size}
% \label{tab:enc_size}
% \end{table*}

\begin{table}
\centering
\small
% \resizebox{\columnwidth}{!}{%
\begin{tabular}{ccc} 
\toprule
% Model & \begin{tabular}[c]{@{}c@{}}Small Tasks\\\small{(Avg.)}\end{tabular} & \begin{tabular}[c]{@{}c@{}}All tasks\\\small{(Avg.)}\end{tabular} \\ 
Model & Small Tasks & All Tasks \\ 
\midrule\midrule
\multicolumn{3}{c}{\encodername} \\ 
\midrule
$\singletask$ & 72.53 & 80.93 \\
$\msparsemtl$ & \textbf{75.88} & \textbf{82.53} \\ 
\midrule\midrule
% \multicolumn{3}{c}{$\mbox{BERT}_{Base}$} \\ 
\multicolumn{3}{c}{\bertbase} \\ 
\midrule
$\singletask$ & 76.53 & 83.34 \\
$\msparsemtl$ & \textbf{80.73} & \textbf{85.45} \\ 
\midrule\midrule
% \multicolumn{3}{c}{$\mbox{BERT}_{Large}$} \\ 
\multicolumn{3}{c}{\bertlarge} \\ 
\midrule
$\singletask$ & 78.85 & 84.93 \\
$\msparsemtl$ & \textbf{82.73} & \textbf{86.94} \\ 
\midrule
$\mbox{MT-DNN}$ & 81.25 & 86.04 \\
\bottomrule
\end{tabular}
% }
\vspace{-0.5em}
\caption{Performance of models with different encoder sizes. \msparsemtl shows consistent gains across encoders of different sizes. \msparsemtl also outperforms the dense MTL baseline MT-DNN~\citep{Liu2019MultiTaskDN}.}
\label{tab:enc_size}
\vspace{-0.5em}
\end{table}
%%%%%%%%%%%%%%%%%%%%%%%%%%%%%%%%%%%%%%
%%%%%%%%%%%%%%%%%%%%%%%%%%%%%%%%%%%%%%
%%%% Task scaling section %%%%
\subsection{Number of Tasks Scaling}
\label{ssec:task_scaling_results}
In this experiment, we evaluate if \msparsemtl can continue to leverage similarities between tasks in the presence of a large number of tasks in its multi-task mixture. To this end, we expand our GLUE multi-task mixture to $16$ tasks with the addition of NLI datasets such as CB; QA datasets such as COPA, MultiRC, and BoolQ; Sentiment datasets such as IMDB, Rotten Tomatoes, and Yelp Polarity; and Word-sense disambiguation datasets such as WiC. For simplicity, we refer to this multitask mixture as \sixteentasks. We provide more details about these datasets in Appendix \ref{app:datasets} and Table~\ref{tab:all_dataset_stats}. We train and evaluate \msparsemtl on this dataset using \bertlarge encoder, and compare with corresponding \singletask baselines on aggregate average performance metrics, \emph{Small Tasks} and \emph{All Tasks}. For \sixteentasks, \emph{Small Tasks} includes RTE, MRPC, STS-B, COLA, Rotten Tomatoes, WiC, CB, BoolQ, and COPA. Table~\ref{tab:task_scaling} shows that \msparsemtl obtains impressive gains, demonstrating the model's ability in scaling to a large number of diverse tasks.

%TODO
\begin{table}
\centering
\small
% \resizebox{\columnwidth}{!}{%
\begin{tabular}{ccc} 
\toprule
\begin{tabular}[c]{@{}c@{}}Model\\\end{tabular} & Small Tasks & All Tasks \\ 
\midrule\midrule
\singletask & 81.28 & 85.00 \\
\msparsemtl & \textbf{83.56} & \textbf{86.46} \\
\bottomrule
\end{tabular}
% }
\vspace{-0.5em}
\caption{Performance comparison on \sixteentasks using \bertlarge. \msparsemtl demonstrates impressive gains on scaling to a large number of diverse tasks.}
\label{tab:task_scaling}
\vspace{-1.5em}
\end{table}

% \footnote{MultiRC is excluded from this list because of its unique formulation -- more details in the Appendix}
% We provide task-level results in Table~\ref{tab:full_16tasks} of Appendix~\ref{app:full_16tasks}

\section{Related Work}
\label{sec:related}
%% Related Work %%
Mixture-of-Experts models have recently achieved promising results by introducing an outrageously large number of parameters while keeping a fixed computation cost via gating mechanism.~\citealp{Shazeer2017OutrageouslyLN} first proposed the MoE layer with a single gating network with $Top$-$k$ routing and load balancing across experts.~\citealp{Fedus2021SwitchTS} propose initialization and training schemes for $Top$-$1$ routing.%~\citealp{Zuo2021TamingSA} propose a consistency regularizer loss for random routing;
~\citealp{yang2021exploring} propose $k$ $Top$-$1$ routing with expert-prototypes, and~\citealp{Roller2021HashLF,Lewis2021BASELS} address other load balancing issues. All the above works study sparse MoE with pre-training from scratch in single-task settings. In contrast, we study multi-task adaptation of such sparse models and devise task-aware gating networks to support MTL. A contemporary work 
%Compared to these works, we propose gating mechanisms geared towards addressing the unique challenges of multi-task learning. Similar to us,
~\cite{Kudugunta2021BeyondDT} studies routing for multi-task training for machine translation, where they route \emph{all} tokens from a task to the same experts with a shared gate. In contrast, we study multi-task {adaptation} where we make routing decisions at token-level using task-specific gates. In the non-Transformer space, an earlier work~\citealp{Ma2018ModelingTR} studied MTL for tabular classification and content recommendation. In contrast to all above works, we study multi-task adaptation of sparse MoE and analyze its robustness for diverse NLU tasks.

%~\citealp{Ma2018ModelingTR} explored a similar idea as ours for FFNs and experimented with multi-task mixtures containing $2$ related tasks. We instead incorporate task-aware gating in sparse Transformers, and demonstrate our approach to be robust to interference across three key dimensions, and to work well across encoder sizes and scale to a large number of diverse NLU tasks.

Multi-task learning and adaptation has been studied extensively for dense models~\cite{caruana1997multitask, Crawshaw2020MultiTaskLW}, with recent works like UnifiedQA~\cite{khashabi-etal-2020-unifiedqa}, MT-DNN~\cite{Liu2019MultiTaskDN} and Muppet~\cite{Aghajanyan2021MuppetMM} showing impressive transfer and low-resource generalization ability. MT-DNN with BERT encoder performs multi-task adaptation on a mixture of GLUE tasks and is used as our baseline. While Muppet also follows similar principles, it uses RoBERTa and much larger number of tasks ($50$). For a fair comparison, with limited compute, we only compare against MT-DNN with the same encoder and same set of MTL tasks. We contrast our MTL setup against the above dense MTL models and demonstrate our sparse design to be more robust on three key transferability aspects.

\section{Conclusion}
\label{sec:conclusion}
%\sg{TODO:}
In this work, we studied multi-task adaptation of sparse MoE models on diverse NLU tasks when initialized with the weights of a pre-trained language model. To support multi-task learning with sparse MoE, we devised task-aware gating networks to route input tokens from different tasks to specialized experts conditioned on the task. We demonstrated such sparse design to be more robust multi-task learners than their non-MOE dense counterparts on several key dimensions including transferability, sample-efficient generalizability, and avoiding catastrophic forgetting.

\section*{Ethical Considerations and Broader Impact}
\label{sec:ethics}
In this work, we develop an efficient multi-task deep neural network model that performs well across several diverse natural language understanding tasks. One of the benefits of a multi-task model is parameter efficiency, where the same model can be used across several different tasks, thereby, saving storage cost and memory footprint. We also demonstrate improved robustness of the multi-task model that further reduces risks of deploying such models in the wild. Furthermore, improved generalization, transferability and sample-efficiency of our model is beneficial for sensitive application domains including finance, legal and healthcare.

However, our model also has the risk of echoing the biases from the pre-trained language model it is based on. Furthermore, a considerable risk with multi-task learning is that it can facilitate the propagation of biases from individual datasets from its training mixture to the rest. Sparse models like \msparsemtl with their increased capability to transfer information from just a single task from its training mixture poses increased risk of retaining and transferring such biases to the unseen tasks. 
Sparse models also massively increase the number of parameters, which can lead to significant storage cost in the absence of customized hardware and optimized implementations, leading to a negative impact on the carbon footprint from training and deploying such models.
% Our experiments and analysis are done on English datasets, and therefore we do not claim that our findings will generalize across all languages, although our framework has potential to be extended to other languages with necessary modifications.

% Furthermore, the multi-task learning of various tasks also has the risk that biases from a dataset from the multi-task mixture might transfer over to other datasets through shared representations. However, since our model supports conditional computation, it has the potential of preventing such cross-task affects by limiting the damage to just a few weights of the network.
% While our framework advance the progress of NLP, it also suffers from associated societal implications of automation ranging from job losses for workers who provide annotations as a service as well as for other industries relying on human labor.

% \section*{Acknowledgements}
% \label{sec:ack}
% \input{ack}

\bibliography{anthology,custom}
\bibliographystyle{acl_natbib}

\clearpage

\appendix

% \clearpage
\section{Appendix}
\label{sec:appendix}
\subsection{Analysis}
\label{app:add_results}
\subsubsection{Re-using task gates for generalization}
\label{ssec:diff_gates}
In Table~\ref{tab:diff_gates}, we provide results for fine-tuning the GLUE-trained \msparsemtl model on unseen SciTail dataset with different task gates. We observe that re-using the gates corresponding to the related tasks (RTE, MNLI) outperforms the random initialization of the gate, as well as re-using the gate from an unrelated task (SST-2). This demonstrates \msparsemtl's ability in learning task-specific routing in its gates, and efficiently re-using it for generalizing to unseen related tasks in a sample efficient manner.
\begin{table}[h]
\centering
\begin{tabular}{cc} 
\toprule
Task Gate & Accuracy \\ 
\midrule\midrule
Random & 91.2 \\ 
\midrule
SST-2 & 91.8 \\
RTE & 92.6 \\ 
\midrule
\textbf{MNLI} & \textbf{92.9} \\
\bottomrule
\end{tabular}
\caption{Performance of \msparsemtl when fine-tuned with different task gates on the 10\% sample of the unseen SciTail dataset. Gates corresponding to tasks with similarity to SciTail (RTE and MNLI) perform superior to random and unrelated task gates (SST-2). All results are with the \encodername encoder.}
\label{tab:diff_gates}
\end{table}

\subsubsection{Task Sampling}
\label{ssec:batching_results}
In Table~\ref{tab:sampling}, we provide results for using different task sampling strategies while training \msparsemtl with heterogeneous batches. We observe that maintaining the natural distributions of tasks during MTL training outperforms uniformly sampling all tasks. We thus use natural sampling of tasks for the MTL models in our experiments.
%Table comparing different Hetero-natural with Hetero-uniform for BERT-Large
\begin{table}[h]
\centering
\begin{tabular}{ccc} 
\toprule
\begin{tabular}[c]{@{}c@{}}Sampling\\\end{tabular} & Small Tasks & All Tasks \\ 
\midrule\midrule
Uniform & 80.60 & 85.75 \\
\textbf{Natural} & \textbf{82.73} & \textbf{86.94} \\
\bottomrule
\end{tabular}
\caption{Comparison of task sampling strategies in \msparsemtl with the \bertlarge encoder on GLUE. Maintaining the natural distribution of tasks (\emph{Natural Sampling}) outperforms uniformly sampling tasks (\emph{Uniform Sampling}).}
\label{tab:sampling}
\end{table}

\subsubsection{Number of Experts}
\label{ssec:num_experts}
In Table~\ref{tab:num_experts}, we provide results for using different number of experts in \msparsemtl. We observe $4$ experts to perform the best, and thus use $4$ experts for all sparse model experiments.
%% Table: Showing the trend with change in the number of experts %%
\begin{table}[h]
\centering
\begin{tabular}{ccc} 
\toprule
\#experts & Small Tasks & All Tasks \\ 
\midrule\midrule
2 experts & 80.78 & 85.79 \\
\textbf{4 experts} & \textbf{82.73} & \textbf{86.94} \\
6 experts & 80.60 & 85.76 \\
\bottomrule
\end{tabular}
\caption{\msparsemtl's performance comparison on GLUE with different number of experts (\#experts) using the \bertlarge encoder. $4$ experts performs the best.}
\label{tab:num_experts}
\end{table}
% \subsection{Experts utilization}
% \label{ssec:experts_utilization}
% %% Discussion about expert utilization and how load balancing, routing loss experiments didn't pan out
% \sg{
% <<The fact that MTL training might be forcing the network to use different experts -- entropy in the plots doesn't seem to be that bad>> <<Insights from our modifications not working that well despite having better expert plots>>, <<Interesting bits from switch usage like how performance always tanks with the usage of a task head, and how not using RTE switch tanks the performance, but not using the COLA switch helps>>, <<Anything else interesting from their weights and gradients>>
% }

% \sg{Our early experiments with increasing expert utilization further through a load balancing and routing loss, adding input jitter to the switch input, and differentiating the expert weights by adding a random noise during initialization didn't work as well.}
\subsection{Datasets}
\label{app:datasets}
%We use a mix of datasets from the GLUE~\cite{Wang2018GLUEAM} and SuperGLUE~\cite{Wang2019SuperGLUEAS} public benchmarks, and augment them with a few additional datasets.
Below, we provide details about all the datasets that we used. We also summarize the key information about these datasets in Table~\ref{tab:all_dataset_stats}.

\noindent{\textbf{RTE}}: Recognizing Textual Entailment are datasets collected from a series of annual textual entailment challenges. The authors combine the data from RTE1~\cite{dagan2006first}, RTE2~\cite{barhaim2006second}, RTE3~\cite{giampiccolo2007third}, and RTE5~\cite{bentivogli2009fifth}. All datasets are converted to two-class classification: entailment and not entailment.

\noindent{\textbf{MRPC}}: Microsoft Research Paraphrase Corpus~\cite{Dolan2005AutomaticallyCA} is a corpus of sentence pairs automatically extracted from online news sources, with human annotations for whether the sentences in the pair are semantically equivalent.

\noindent{\textbf{STS-B}}: Semantic Textual Similarity Benchmark~\cite{Cer2017SemEval2017T1} is a collection of sentence pairs drawn from news headlines, video and image captions, and natural language inference data. Each pair is human-annotated with a similarity score from 1 to 5.

\noindent{\textbf{CoLA}}: Corpus of Linguistic Acceptability~\cite{Warstadt2019NeuralNA} consists of English acceptability judgments drawn from books and journal articles on linguistic theory. Each example is a sequence of words annotated with whether it is a grammatical English sentence.

\noindent{\textbf{SST-2}}: Stanford Sentiment Treebank~\cite{Socher2013RecursiveDM} consists of sentences from movie reviews and human annotations of their sentiment. The task is to predict the sentiment of a given sentence. It uses the two-way (positive/negative) class split, with only sentence-level labels.

\noindent{\textbf{QNLI}}: Stanford Question Answering Dataset~\citep{Wang2018GLUEAM, Rajpurkar2016SQuAD1Q} is a question-answering dataset consisting of question-paragraph pairs, where one of the sentences in the paragraph (drawn from Wikipedia) contains the answer to the corresponding question (written by an annotator). The authors of the benchmark convert the task into sentence pair classification by forming a pair between each question and each sentence in the corresponding context, and filtering out pairs with low lexical overlap between the question and the context sentence. The task is to determine whether the context sentence contains the answer to the question. This modified version of the original task removes the requirement that the model select the exact answer, but also removes the simplifying assumptions that the answer is always present in the input and that lexical overlap is a reliable cue.

\noindent{\textbf{QQP}}: Quora Question Pairs2 dataset~\cite{iyerqqp} is a collection of question pairs from the community question-answering website Quora. The task is to determine whether a pair of questions are semantically equivalent.

\noindent{\textbf{MNLI}}: Multi-Genre Natural Language Inference Corpus~\cite{Williams2018ABC} is a crowdsourced collection of sentence pairs with textual entailment annotations. Given a premise sentence and a hypothesis sentence, the task is to predict whether the premise entails the hypothesis (entailment), contradicts the hypothesis (contradiction), or neither (neutral). The premise sentences are gathered from ten different sources, including transcribed speech, fiction, and government reports. The authors of the benchmark use the standard test set, for which they obtained private labels from the RTE authors, and evaluate on both the matched (in-domain) and mismatched (cross-domain) section.

\noindent{\textbf{CB}}: Commitment Bank~\cite{de2019commitmentbank} is a corpus of short texts in which at least one sentence contains an embedded clause. Each of these embedded clauses is annotated with the degree to which it appears the person who wrote the text is committed to the truth of the clause. The resulting task framed as three-class textual entailment on examples that are drawn from the Wall Street Journal, fiction from the British National Corpus, and Switchboard. Each example consists of a premise containing an embedded clause and the corresponding hypothesis is the extraction of that clause.

\noindent{\textbf{BoolQ}}: Boolean Questions~\cite{Clark2019BoolQET} is a QA task where each example consists of a short passage and a yes/no question about the passage. The questions are provided anonymously and unsolicited by users of the Google search engine, and afterwards paired with a paragraph from a Wikipedia article containing the answer.

\noindent{\textbf{MultiRC}}: Multi-Sentence Reading Comprehension~\cite{Khashabi2018LookingBT} is a QA task where each example consists of a context paragraph, a question about that paragraph, and a list of possible answers. The system must predict which answers are true and which are false. Each question can have multiple possible correct answers, so each question-answer pair must be evaluated independent of
other pairs. The questions are also designed such that answering each question requires drawing facts
from multiple context sentences. The
paragraphs are drawn from seven domains including news, fiction, and historical text.

\noindent{\textbf{WiC}}: Word-in-Context~\cite{Pilehvar2019WiCTW} is a word sense disambiguation task cast as binary classification of sentence pairs. Given two text snippets and a polysemous word that appears in both sentences, the task is to determine whether the word is used with the same sense in both sentences.

\noindent{\textbf{COPA}}: Choice of Plausible Alternatives~\cite{roemmele2011choice} is a causal reasoning task in which a system is given a premise sentence and must determine either the cause or effect of the premise from two possible choices. All examples are handcrafted and focus on topics from blogs and a photography-related encyclopedia.

\noindent{\textbf{IMDB}}: Large Movie Review Dataset ~\cite{Maas2011LearningWV} built from reviews from IMDb (Internet Movie Database). This is a dataset for binary sentiment classification containing highly polar movie reviews.

\noindent{\textbf{Yelp Polarity}}: Large Yelp Review Dataset~\cite{Zhang2015CharacterlevelCN}. This is a dataset for binary sentiment classification constructed from highly polar Yelp reviews.

\noindent{\textbf{Rotten Tomatoes}}: Movie Review Dataset~\cite{Pang2005SeeingSE}. This is a dataset of containing positive and negative processed sentences from Rotten Tomatoes movie reviews.

\begin{table*}[ht!]
\centering
\resizebox{!}{0.47\textheight}{
\begin{tabular}{cccccc}
\toprule
Dataset & \#Train & \#Dev & \#Labels & Formulation & Metrics \\ 
\midrule\midrule
\multicolumn{6}{c}{\textbf{WSD}} \\ 
\midrule
WiC & 5.4k & 638 & 2 & \begin{tabular}[c]{@{}c@{}}Pairwise-text\\Classification\end{tabular} & Accuracy \\ 
\midrule\midrule
\multicolumn{6}{c}{\textbf{Similarity}} \\ 
\midrule
STS-B & 5.7k & 1.5k & 1 & \begin{tabular}[c]{@{}c@{}}Pairwise-text\\Regression\end{tabular} & Spearman corr \\ 
\midrule\midrule
\multicolumn{6}{c}{\textbf{Acceptability}} \\ 
\midrule
CoLA & 8.5k & 1k & 2 & \begin{tabular}[c]{@{}c@{}}Single-text\\Classification\end{tabular} & Matthews corr \\ 
\midrule\midrule
\multicolumn{6}{c}{\textbf{Sentiment}} \\ 
\midrule
\begin{tabular}[c]{@{}c@{}}Rotten \\Tomatoes\end{tabular} & 8.5k & 1k & 2 & \begin{tabular}[c]{@{}c@{}}Single-text\\Classification\end{tabular} & Accuracy \\ 
\midrule
IMDB & 25k & 25k & 2 & \begin{tabular}[c]{@{}c@{}}Single-text\\Classification\end{tabular} & Accuracy \\ 
\midrule
SST-2 & 67.3k & 872 & 2 & \multicolumn{1}{c}{\begin{tabular}[c]{@{}c@{}}Single-text\\Classification\end{tabular}} & Accuracy \\ 
\midrule
\begin{tabular}[c]{@{}c@{}}Yelp\\Polarity\end{tabular} & 560k & 38k & 2 & \begin{tabular}[c]{@{}c@{}}Single-text\\Classification\end{tabular} & Accuracy \\ 
\midrule\midrule
\multicolumn{6}{c}{\textbf{Paraphrase}} \\ 
\midrule
MRPC & 3.7k & 408 & 2 & \begin{tabular}[c]{@{}c@{}}Pairwise-text\\Classification\end{tabular} & Accuracy \\ 
\midrule
QQP & 364k & 40k & 2 & \begin{tabular}[c]{@{}c@{}}Pairwise-text\\Classification\end{tabular} & Accuracy \\ 
\midrule\midrule
\multicolumn{6}{c}{\textbf{NLI}} \\ 
\midrule
CB & 250 & 56 & 3 & \begin{tabular}[c]{@{}c@{}}Pairwise-text\\Classification\end{tabular} & Accuracy \\ 
\midrule
RTE & 2.5k & 277 & 2 & \begin{tabular}[c]{@{}c@{}}Pairwise-text\\Classification\end{tabular} & Accuracy \\ 
\midrule
SciTail & 23.6k & 1.3k & 2 & \begin{tabular}[c]{@{}c@{}}Pairwise-text\\Classification\end{tabular} & Accuracy \\ 
\midrule
QNLI & 105k & 5.5k & 2 & \begin{tabular}[c]{@{}c@{}}Pairwise-text\\Classification\end{tabular} & Accuracy \\ 
\midrule
MNLI & 393k & 9.8k & 3 & \begin{tabular}[c]{@{}c@{}}Pairwise-text\\Classification\end{tabular} & Accuracy \\ 
\midrule\midrule
\multicolumn{6}{c}{\textbf{QA}} \\ 
\midrule
COPA & 400 & 100 & 2 & \begin{tabular}[c]{@{}c@{}}Pairwise-text\\Ranking\end{tabular} & Accuracy \\ 
\midrule
MultiRC & 27k & 4.8k & 2 & \begin{tabular}[c]{@{}c@{}}Pairwise-text\\Classification\end{tabular} & F1a \\ 
\midrule
BoolQ & 9.4k & 3.3k & 2 & \begin{tabular}[c]{@{}c@{}}Pairwise-text\\Classification\end{tabular} & Accuracy \\
\bottomrule
\end{tabular}
}
\caption{Key information about all the datasets used.}
\label{tab:all_dataset_stats}
\end{table*}

\noindent{\textbf{SciTail}}: SciTail~\cite{Khot2018SciTaiLAT} dataset is an entailment dataset created from multiple-choice science exams and web sentences. Each question and the correct answer choice are converted into an assertive statement to form the hypothesis. Information retrieval is used to obtain relevant text from a large text corpus of web sentences, and use these sentences as a premise. Premise-hypothesis pair are annotated as supports (entails) or not (neutral).

We obtained all of these datasets from HuggingFace's datasets library~\cite{Lhoest2021DatasetsAC}.

% \input{tables/glue_dataset_stats}
% \subsection{Pseudocode for MTL training}
% \label{app:pseudocode}
\subsection{Implementation Details}
\label{app:implementation}
\subsubsection{Task formulations}
\label{app:task_formulations}
In this section, we group all the tasks into different categories, and provide details about their formulation. All model variants followed BERT-like architectures~\cite{Devlin2019BERTPO} with a [CLS] token added to the beginning of the input.
\subsubsection*{Single-text Classification}
CoLA, SST-2, IMDB, Yelp Polarity, and Rotten Tomatoes belong to this category. The task is to perform binary classification based on a single sequence of concatenated sentences.  A classifier head is used on top of the output representation of the [CLS] token for the classification. We use Matthews correlation coefficient~\cite{MatthewsCorr1975} as the evaluation metric for CoLA, and use accuracy for the rest.
\subsubsection*{Pairwise-text Classification}
RTE, MRPC, QNLI, QQP, MNLI, CB, BoolQ, MultiRC, WiC, and SciTail belong to this category. The task is to perform binary or multi-class classification based on a pair of sequence inputs. We concatenate the input sequence pairs separated by a [SEP] token following~\cite{Devlin2019BERTPO}, and feed the fused sequence to the model. In the case of \mbox{MultiRC}, which contains three sequences (paragraph, question, and answer), the paragraph and question are concatenated to form the first sequence, and the answer is used as the second sequence. For all tasks except WiC, a classifier head which sees the output representation corresponding to the [CLS] token is used to select the predicted class. For WiC a span classification head is used, which extracts the output representations of the word of interest (from both input sentences) and concatenates them with the representation of the [CLS] token. This fused representation is then fed to a classifier head to predict the binary output. Following the authors, we use F$1_a$ as the metric for MultiRC, which evaluates binary decisions on all the answer-options in the dataset independently. F$1_a$ is the harmonic mean of precision and recall across all answer-option pairs, without grouping by question or paragraph. For all other tasks, accuracy is used as the evaluation metric.
\subsubsection*{Pairwise-text Ranking}
COPA belongs to this category. The task is to choose between a pair of sequences given a premise-question context. We join the premise-question sequence pair into a single context sequence, and evaluate each pair of choice alternatives independently by concatenating context, [SEP] token, and answer choice to form a pair of input sequences for the model. The task is then cast as a binary classification task for each input pair, for which we feed the output representation to a classifier head, and retrieve the positive (True) class logits for each input. Whichever input returns the largest positive-class logit is then taken as the answer choice, and we calculate accuracy as the evaluation metric.
\subsubsection*{Pairwise-text Regression}
STS-B belongs to this category. The task is to perform regression from a pair of input sequences. The input sequences are concatenated together with a [SEP] token and fed to the model. A regression head is used to learn the similarity score and we calculate Spearman's rank correlation as the evaluation metric.

\subsubsection{Model details}
We use a Wordpiece Tokenizer~\cite{Wu2016GooglesNM} with $30$k vocabulary size to tokenize all the examples. We truncate the examples on the right using a maximum length of $512$ for QNLI and MNLI, and $128$ for the rest of GLUE datasets. We use a batch size of $128$ for MTL Training, and $32$ for fine-tuning. %We re-use the task-heads and gates during fine-tuning on GLUE and GLUE++.

For training of Sparse models, we do not add any additional load balancing loss, input jitter, or additional dropout in the experts\footnote{Early experiments resulted in a drop in the performance.}. \emph{Unlike existing work, we did not encounter a load-imbalance in the utilization of the experts, potentially due to the multi-task objective that pushes the network to specialize weights in different experts.}

\subsection*{Model selection}
\label{app:model_selection}
For MTL training, we train the model for a fixed number of steps, and select the checkpoint at the end of training. For fine-tuning, we use early stopping using the dev set. We tune the learning rate, warmup proportion, and the number of training steps for both MTL Training and fine-tuning. For fine-tuning, tuning is only done for the small tasks ($<10k$ examples)\footnote{Bigger tasks showed indifference to the choice of hyperparameters.}. For every task, we run $3$ fine-tuning experimental runs for each model with different seeds, and report the max number obtained across runs for the model.

\subsection*{Hyper-parameters and Tuning}
\label{app:tuning}
For the Adam optimizer, we used $\beta_1$ and $\beta_2$ values of 0.9 and 0.999 respectively, and an $\epsilon$ of $1e-8$. For MTL Training, we ran tuning runs with a grid search of the learning rate in $[5e-06, 1e-05, 2e-05, 5e-05, 1e-04]$, warmup rate in $[0.1, 0.2]$, and number of steps in $[30k, 50k]$. For fine-tuning, we tuned the learning rate in $[5e-06, 1e-05, 2e-05, 5e-05, 1e-04]$, used a warmup of $0.1$, and tuned the number of epochs in $[5, 10, 15, 20, 25, 30]$.
%and used the random seed values in $[42, 2048, 5]$.

%For the OneCycleLR schedule, we use a warmup of 0.1, a div\_factor of 1e7, a final\_div\_factor of 1e10, and a linear anneal\_strategy.\\
\subsection{Limitations and Future Work}
\label{app:limitations}
Using a separate gate for each task allows us to learn task-specific routing in the gates, however, it has the limitation that individual gates are only updated via the examples corresponding to their target task. This can lead to the gates for the smallest tasks being under-trained under a natural sampling of tasks. In the future, we will experiment with a training schedule in which we use uniform sampling at the beginning of training to allow all gates to train sufficiently, and then revert back to natural sampling. Our method also has the limitation that gates of related tasks only share information via the experts. To tackle this, we will experiment with incorporating task embeddings to allow the network to share routing information by learning similar task embeddings for related tasks. Lastly, we will experiment with further scaling up the number and diversity of tasks in our multitask mixture to obtain a general model for a wide-range of downstream tasks.

\begin{table*}[t]
\centering
\small
\begin{tabular}{ccccccccc} 
\toprule
Dataset & \multicolumn{1}{c}{\begin{tabular}[c]{@{}c@{}}RTE\\\small{(2.5k)}\end{tabular}} & \begin{tabular}[c]{@{}c@{}}MRPC\\\small{(3.7k)}\end{tabular} & \begin{tabular}[c]{@{}c@{}}STS-B\\\small{(5.7k)}\end{tabular} & \begin{tabular}[c]{@{}c@{}}QNLI\\\small{(105k)}\end{tabular} & \begin{tabular}[c]{@{}c@{}}QQP\\\small{(364k)}\end{tabular} & \multicolumn{1}{c}{\begin{tabular}[c]{@{}c@{}}MNLI\\\small{(393k)}\end{tabular}} & \begin{tabular}[c]{@{}c@{}}Small Tasks\\\small{(Avg.)}\end{tabular} & \multicolumn{1}{c}{\begin{tabular}[c]{@{}c@{}}All Tasks\\\small{(Avg.)}\end{tabular}} \\ 
\midrule\midrule
\multicolumn{9}{c}{\densemtl} \\
\midrule
\sixtasks & \textbf{78.6} & \textbf{89.7} & 90.5 & 89.8 & \textbf{90.9} & 83.6 & \textbf{86.27} & \textbf{87.18} \\ 
\eighttasks & 77.9 & 89 & 90.5 & \textbf{90.3} & 90.8 & \textbf{83.8} & 85.80 & 87.05 \\ 
\midrule\midrule
\multicolumn{9}{c}{\sparsemtl} \\
\midrule
\sixtasks & 78.9 & 89.5 & 90.4 & 90.1 & 90.9 & 83.5 & 86.27 & 87.22 \\ 
\eighttasks & 78.9 & \textbf{90} & \textbf{90.5} & \textbf{90.3} & 90.9 & \textbf{83.6 } & \textbf{86.47} & \textbf{87.37} \\ 
\midrule\midrule
\multicolumn{9}{c}{\textbf{\msparsemtl}} \\
\midrule
\sixtasks & 78.2 & \textbf{90.9} & 90.4 & 90 & 90.8 & 83.6 & 86.50 & 87.32 \\ 
\eighttasks & \textbf{81.1} & 90.7 & \textbf{90.6} & \textbf{90.2 } & 90.8 & 83.6 & \textbf{87.47} & \textbf{87.83} \\
\bottomrule
\end{tabular}
\caption{Task-level model performance on GLUE (containing several diverse tasks) and \sixtasks (as a subset of GLUE containing only related tasks) evaluated on the common tasks in both. Sparse MTL models demonstrate robustness in the presence of unrelated tasks in GLUE, with \msparsemtl with task-specific routing being the most robust. All models use MiniLM encoder.}
\label{tab:full_robustness}
\end{table*}
\subsection{Task-level Results}
\subsubsection{Robustness to unrelated tasks}
\label{app:full_robustness}
We provide the task-level results corresponding to the robustness experiments from Section~\ref{ssec:robustness_results} in Table~\ref{tab:full_robustness}.

\begin{table*}[t]
\centering
\small
\resizebox{\linewidth}{!}{%
\begin{tabular}{ccccccccccc} 
\toprule
Model & \begin{tabular}[c]{@{}c@{}}RTE\\\small{(2.5k)}\end{tabular} & \begin{tabular}[c]{@{}c@{}}MRPC\\\small{(3.7k)}\end{tabular} & \begin{tabular}[c]{@{}c@{}}STS-B\\\small{(5.7k)}\end{tabular} & \begin{tabular}[c]{@{}c@{}}CoLA\\\small{(8.5k)}\end{tabular} & \begin{tabular}[c]{@{}c@{}}SST-2\\\small{(67.3k)}\end{tabular} & \begin{tabular}[c]{@{}c@{}}QNLI\\\small{(105k)}\end{tabular} & \begin{tabular}[c]{@{}c@{}}QQP\\\small{(364k)}\end{tabular} & \begin{tabular}[c]{@{}c@{}}MNLI\\\small{(393k)}\end{tabular} & \begin{tabular}[c]{@{}c@{}}Small Tasks\\\small{(Avg.)}\end{tabular} & \begin{tabular}[c]{@{}c@{}}All tasks\\\small{(Avg.)}\end{tabular} \\ 
\midrule\midrule
\multicolumn{11}{c}{$\mbox{BERT}_{Base}$} \\ 
\midrule\midrule
$\singletask$ & 71.4 & 84.8 & 89.1 & \uline{60.8} & \textbf{92.9} & \textbf{91.9} & \textbf{91.4} & 84.4 & 76.53 & 83.34 \\
$\msparsemtl$ & \textbf{81.1} & \textbf{90.7} & \textbf{90.4} & \textbf{60.7} & \textbf{92.9} & 91.8 & \textbf{91.4} & 84.6 & \textbf{80.73} & \textbf{85.45} \\ 
\midrule\midrule
\multicolumn{11}{c}{$\mbox{BERT}_{Large}$} \\ 
\midrule\midrule
$\singletask$ & 74.6 & 88.2 & 89.9 & 62.7 & 93.3 & \uline{92.7} & \textbf{91.7} & 86.3 & 78.85 & 84.93 \\
$\msparsemtl$ & \textbf{86.4} & \textbf{89.2} & \textbf{90.8} & \textbf{64.5} & \uline{94.2} & 92.3 & \textbf{91.7} & \uline{86.4} & \textbf{82.73} & \textbf{86.94} \\ 
\midrule
$\mbox{MT-DNN}$ & 83.4 & 87.5 & 90.6 & 63.5 & \textbf{94.3} & \textbf{92.9} & 89.2 & \textbf{86.9} & 81.25 & 86.04 \\
\bottomrule
\end{tabular}
}
\caption{Task-level performance of models with different BERT encoder sizes. \msparsemtl shows consistent gains across encoders of different sizes. \msparsemtl also outperforms the dense MTL baseline MT-DNN~\citep{Liu2019MultiTaskDN}.}
\label{tab:full_enc_size}
\end{table*}
\subsubsection{Encoder Scaling}
\label{app:full_encoder_size}
We provide the task-level results corresponding to the encoder scaling experiments from Section~\ref{ssec:encoder_size_results} in Table~\ref{tab:full_enc_size}.
% \input{tables/full_16tasks}
% \subsubsection*{Task Scaling}
% \label{app:full_16_tasks}

\end{document}